\pdfoutput=1
\documentclass[10pt,twocolumn,letterpaper]{article}

\usepackage{acl}

\usepackage{graphicx}
\usepackage{amsmath}
\usepackage{amssymb}
\usepackage{booktabs}

\usepackage{times}
\usepackage{latexsym}
\usepackage{adjustbox}
\usepackage{listings}
\usepackage[T1]{fontenc}
\usepackage[utf8]{inputenc}
\usepackage{microtype}
\usepackage{inconsolata}
\usepackage{graphicx}
\usepackage{multirow}
\usepackage{booktabs}
\usepackage{amssymb}
\usepackage{pifont}
\usepackage{fvextra}

\usepackage[capitalize]{cleveref}
\crefname{section}{Sec.}{Secs.}
\Crefname{section}{Section}{Sections}
\Crefname{table}{Table}{Tables}
\crefname{table}{Tab.}{Tabs.}

\newcommand{\cmark}{\checkmark}
\newcommand{\xmark}{\ding{55}}

\newcommand{\datasetname}{HinTel-AlignBench}
\newcommand{\vqasource}{VQAv2}
\newcommand{\realworldqa}{RealWorldQA}
\newcommand{\clevrmath}{CLEVR-Math}

\newcommand{\jee}{JEE}

\DefineVerbatimEnvironment{PromptBlock}{Verbatim}{
  breaklines,
  breakanywhere=true,
  fontsize=\small,
  commandchars=\\\{\},
  breaksymbolleft={}
}

\begin{document}

\title{\datasetname: A Framework and Benchmark for Hindi–Telugu with English-Aligned Samples}


\author{
\\
  \textbf{Rishikant Chigrupaatii}\textsuperscript{1}\thanks{Equal contribution} \quad
  \textbf{Ponnada Sai Tulasi Kanishka}\textsuperscript{1,*} \quad
  \textbf{Lalit Chandra Routhu}\textsuperscript{1,*} \quad
  \textbf{Martin Patel}\textsuperscript{1,*} \\[0.5ex] 
  \textbf{Sama Supratheek Reddy}\textsuperscript{1} \quad
  \textbf{Divyam Gupta}\textsuperscript{1} \quad
  \textbf{Dasari Srikar}\textsuperscript{1} \quad
  \textbf{Krishna Teja Kuchimanchi}\textsuperscript{1} \\[0.5ex]
  \textbf{Rajiv Misra}\textsuperscript{1} \quad
  \textbf{Rohun Tripathi}\textsuperscript{2} \\
  \\
  \textsuperscript{1}Indian Institute of Technology Patna \quad \textsuperscript{2}Allen Institute for AI \\
  {\tt\small \{rishikant\_2101cs66, ponnada\_2101cs57, lalit\_2101ai17, martin\_2101cs43\}@iitp.ac.in} \\
}
\setlength\titlebox{5cm}

\maketitle

\begin{abstract}
With nearly 1.5 billion people and more than 120 major languages, India represents one of the most diverse regions in the world. As multilingual VLMs gain prominence, robust evaluation methodologies are essential to drive progress toward equitable AI for low-resource languages. Current multilingual VLM evaluations suffer from four major limitations: reliance on unverified auto-translations, narrow task / domain coverage, limited sample sizes, and lack of cultural and natively sourced QA. To address these gaps, we present a scalable framework to evaluate VLMs in Indian languages and compare it with performance in English. Using the framework, we generate HinTel-AlignBench\footnote{\url{https://rishikant24.github.io/indicvisionbench.github.io/}}, a benchmark that draws from diverse sources in Hindi and Telugu with English aligned samples. Our contributions are threefold: (1) a semi-automated dataset creation framework combining back-translation, filtering, and human verification; (2) the most comprehensive vision-language benchmark for Hindi and Telugu, including adapted English datasets (VQAv2, RealWorldQA, CLEVR-Math) and native novel Indic datasets (JEE for STEM, VAANI for cultural grounding) with $\sim4k$ QA per language; and (3) a detailed performance analysis of various SOTA open weight and closed source VLMs. We find a regression in performance for tasks in English vs. in Indian languages for 4 out of 5 tasks across all the models, with an average regression of 8.3 points in Hindi and 5.5 for Telugu. We categorize common failure modes to highlight concrete areas of improvement in multilingual multimodal understanding.
\end{abstract}

\section{Introduction}
\label{sec:intro}

\begin{figure}[ht]
    \centering
    \includegraphics[width=\linewidth]{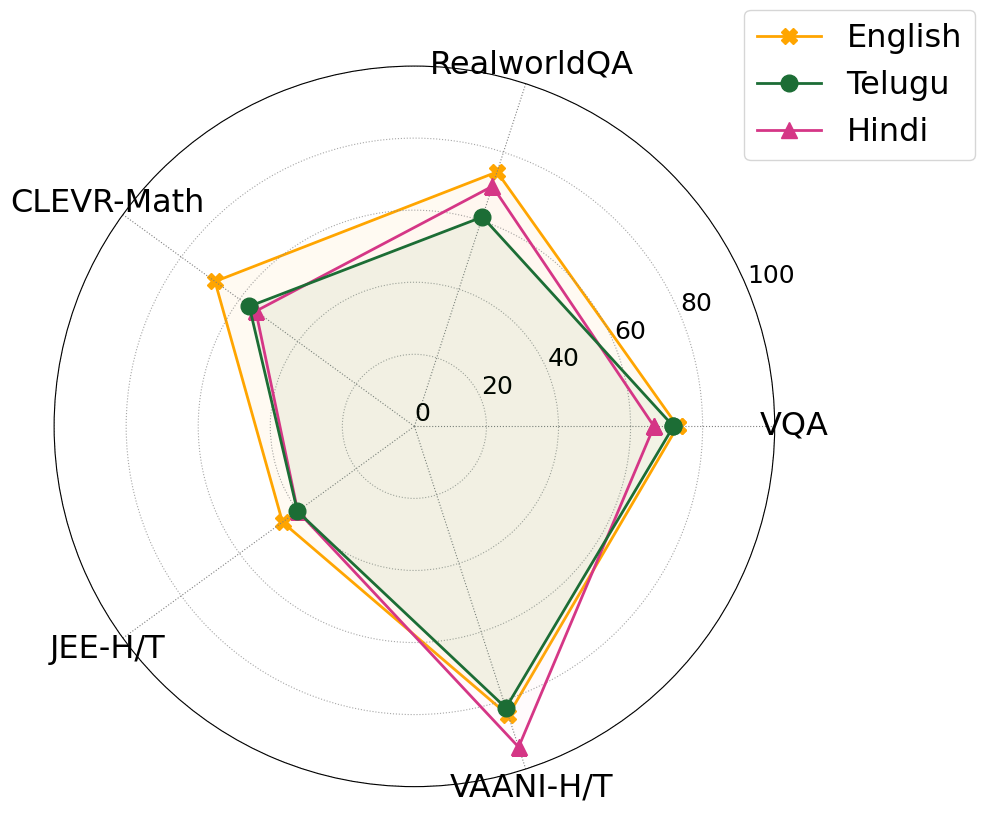} 
    \caption{Average Performance of GPT-4.1 and Gemini-2.5-Flash models on English, Hindi and Telugu Languages on data-parallel visual question answering samples. Overall, performance regresses from English to Hindi by 8.3 points and regresses from English to Telugu by 5.5 points.}
    \label{fig:teaser-star-map}
\end{figure}

India is home to an extraordinary diversity of languages, with 122 major languages and 1599 other languages\footnote{\url{https://en.wikipedia.org/wiki/Languages_of_India}}. Despite this linguistic diversity, most state-of-the-art Multimodal Large Language Models (MLLMs) have historically been English-centric or focused on high-resource European languages. There has been a recent surge in multilingual MLLMs that support multiple languages, including frontier models~\cite{openai2025gpt41, google2025gemini25, meta2025llama32vision, anthropic2025claude37, 2025ayavision}. However, there is no unified benchmark per Indian language to evaluate these models. Various benchmarks have been proposed for evaluating multilingual LLMs~\cite{ahuja2023megamultilingualevaluationgenerative, singh-etal-2024-indicgenbench} but evaluating MLLMs fairly and comprehensively for Indian and other low-resource languages remains a significant challenge \cite{romero2024cvqa}.

Current evaluation methodologies for multilingual vision-language models suffer from four fundamental limitations. First, many benchmarks rely on automatically translated datasets without human verification e.g., m-WildVision~\cite{2025ayavision}, inevitably introducing noise and artifacts. As noted in \cite{wu2025bitterlessonlearned2000}, auto-translated evaluation sets with and without back-translation often introduce biases, while relying entirely on manual curation is too slow for scalable benchmarking. Second, existing evaluations suffer from narrow scope, both in task diversity and domain coverage \cite{salazar2025kaleidoscopeinlanguageexamsmassively}. This myopic focus fails to assess whether models can handle different domains and tasks ranging from real-world visual question answering to mathematical reasoning and cultural awareness in the target language. Third, commonly used benchmarks have very limited sample sizes: xChat \cite{yue2025pangea} and AyaVisionBench \cite{2025ayavision} have 50 and 135 QA per language, respectively. Fourth, current approaches like BharatBench's \cite{khan2024chitrarth} use of translated LLaVABench \cite{llava-liu2023visual}, MM-Vet \cite{yu2024mmvetevaluatinglargemultimodal} and Pope \cite{li2023evaluating} datasets fundamentally miss cultural grounding, linguistic nativity, and domain diversity, evaluating only surface-level translation quality rather than true multilingual competence. These limitations compound and decrease the reliability of benchmarks to measure progress in the support of Indian languages, ultimately slowing the development of equitable multilingual vision systems.

To address these issues, we introduce an efficient framework for generating high-quality, visually diverse multilingual evaluation sets. The framework employs a translation or text only LLM based QA generation stage followed by human review and verification stage. In the human review, each sample is assessed for semantic accuracy, linguistic style, and readability. Our semi-automated approach accelerates benchmark generation while maintaining rigorous linguistic fidelity through manual review. Our framework uses translation followed by verification, which is an easier annotation task and is significantly faster than generating question answer pairs from scratch, with processing speeds 5x faster for more than 79\% of the samples as compared to generating question and answer pairs.

We follow this framework to create the most comprehensive vision language benchmark for Hindi and Telugu, featuring: (1) Adapted English benchmarks: Carefully adapted versions of real-world visual QA (VQAv2~\cite{balanced_vqa_v2}), practical reasoning (RealWorldQA~\cite{xai2024realworldqa}), and visual mathematical reasoning (CLEVR-Math~\cite{lindstrom2022-clevr-math}) and (2) Native Indic evaluation sets: JEE-Vision: STEM competency testing via India's Joint Entrance Exam problems and VAANI \cite{vaani2025}: Cultural and region-specific visual QA covering traditions, artifacts, and local contexts. Our selected subsets provide a multi-domain coverage and represent a broad assessment of multilingual vision language capabilities. 
For each sample, we also generate and manually verify the English sample. This gives us aligned samples in the target Indian language and English, which then enables an exact comparison of the model across the two languages.
As far as we are aware, these are the most diverse collection of Hindi and Telugu VQA evaluation sets with English-aligned samples.

We comprehensively evaluate various open-weight and frontier closed-source models on our benchmark. 
There is a performance regression of 8.3 points from English to Hindi and 5.5 points from English to Telugu on average in evaluated models, \Cref{fig:teaser-star-map}. Surprisingly, even for frontier SoTA models such as GPT 4.1, we find a 3.8-point drop in performance from English to Hindi and a 8.6-point drop from English to Telugu. Finally, the performance across Hindi and Telugu on aligned subsets is within 1 point, demonstrating a similar performance gap between English and both Indian languages.

By establishing a reliable evaluation methodology, we aim to enable the creation of robust evaluation datasets for low-resource languages worldwide. Our contributions are as follows.
\begin{itemize}
    \item \textbf{A Framework}: A semi-automated, scalable system for generating high-quality multilingual vision language evaluation sets.
    \item \textbf{A Benchmark}: The large-scale, visually-diverse and publicly available evaluation dataset for Hindi and Telugu VLMs to support rigorous and consistent evaluation.
    \item \textbf{Comprehensive Evaluation}: An in-depth analysis of \textbf{SOTA vision LLMs}, highlighting regression in performance from English to Indic language in 4 out 5 domains and up to 15.6 points in the worst case.
\end{itemize}

\begin{table*}[ht]
\centering
\small
\begin{tabular}{l ll l r r c c c c}
\toprule
\textbf{Benchmark} & \textbf{Indic Lang} & \textbf{QA Type} & \shortstack{\textbf{Image} \\ \textbf{count}} & \textbf{QA} & \shortstack{\textbf{Indic QA} \\ \textbf{per lang}} & \textbf{Human} & \shortstack{\textbf{Culturally} \\ \textbf{Sourced}} \\
\midrule
AyaVisionBench & \textbf{hin} & Chat & 3.1k &  3.1k & 135 & \xmark & \xmark \\
xMMMU & \textbf{hin} & MC & 300 & 3k & 291 & \xmark & \xmark \\
xGQA & ben & OE & 300 & 77.3k & 9.7k & \xmark & \xmark\\
MTVQA  & - & OE & 8.8k & 28.6k & - & \cmark & \xmark \\
M3Exam & - & MC & 2.8k & 12.3k & - & \cmark & \xmark \\
xChatBench & \textbf{hin} & Chat & 400 & 400 & 50 & \cmark & \xmark\\
Kaleidoscope & ben, \textbf{hin}, \textbf{tel} & MC & 20.9k & 20.9k & $\sim800$ & \cmark & \xmark\\
XM100 & ben, \textbf{hin}, \textbf{tel} & Caption & 100 & 3.6k & 100 & \cmark& \cmark \\
MaRVL & tam & Caption & 5.5k & 5.7k & 1.2k & \cmark &\cmark\\
MaXM & \textbf{hin} & OE & 1.4k & 2.1k & 294 & \cmark & \cmark \\
CVQA  & ben, urd, tam, mar, \textbf{hin}, \textbf{tel} & MC & 5.2k & 10.4k & $\sim300$ & \cmark& \cmark \\
\midrule
Ours & \textbf{hin}, \textbf{tel} & OE, MC & 5.1k & 13.5k & $\sim4k$ & \cmark & \cmark \\
\bottomrule
\end{tabular}
\caption{Comparison of existing multilingual Visual Question Answering (VQA) benchmarks with ours.%
\textbf{“QA Type”} denotes the question‑answering format %
  (Chat = Multimodal chat; OE = open‑ended VQA;%
   MC = multiple‑choice VQA; Caption = captioning/multi-image captioning); %
\textbf{“QA”} denotes the total question‑answer pairs; %
\textbf{“Indic QA per lang”} denotes the question‑answer pairs per Indic language; %
\textbf{“Human”} indicates human verified/annotated data; %
\textbf{"Culturally Sourced"} indicates culturally-sourced data.
xChatBench, xMMMU and XM100 are part of PangeaBench~\cite{yue2025pangea}.
Ours (bottom row) is the only VQA dataset with human‑verified annotations, diverse culturally sourced samples and a significant sample size per Indic language.}
\label{tab:multilingual_vqa_compact}
\end{table*}

\section{Related Work}

\subsection{Multimodal Instruction-Tuned Models}
Previous work has shown that large language models can be extended to vision tasks by instruction tuning with multimodal data~\cite{llava-liu2023visual, Dai2023InstructBLIPTG, blip2-li2023blip, minigpt-4}.
Many open weight Vision-Language Models such as Gemma3 \cite{team2025gemma}, Qwen2.5VL \cite{bai2025qwen25vl}, InternVL3 \cite{zhu2025internvl3}, Molmo \cite{deitke2024molmo}, Llama3.2 Vision \cite{meta2025llama32vision} and closed source models such as GPT-4.1 \cite{openai2025gpt41}, Claude 3.7 \cite{anthropic2025claude37}, Gemini 2.5 \cite{google2025gemini25}) have since been released, all leveraging some form of multimodal instruction tuning. Many of these recent models are multilingual, but the evaluation of their multilingual abilities is limited and is not reported on a consistent and reliable multimodal multilingual benchmark \cite{2025ayavision, yue2025pangea, palowacv, alam2025behind-maya-vlm}.

\subsection{Multilingual Vision-Language Benchmarks}
There have been a number of recent benchmarks which evaluate multi-lingual abilities and cultural robustness of VLMs. CVQA \cite{romero2024cvqa} contains 10.4k visual QA pairs across 31 languages, demonstrating large performance gaps on low-resource languages. MTVQA \cite{tang2024mtvqa} provides text-centric VQA in 9 languages with human annotations. xGQA \cite{pfeiffer-etal-2022-xgqa} automatically translates the GQA \cite{hudson2018gqa} dataset into 7 languages while MaxM \cite{changpinyo2022maxm} uses Machine Translation plus lightweight post‐editing for 7 languages, with less than 300 samples per language. MarVL \cite{liu-etal-2021-visually-MarVL} focuses on culturally sourced reasoning by evaluating whether a caption about an image pair is true or false, but is not a Visual Question Answering dataset.
Concurrently with our work, Kaleidoscope \cite{salazar2025kaleidoscopeinlanguageexamsmassively} is a multi-lingual multimodal benchmark with ~800 VQA MCQs per Indic language sourced from regional examinations. However, Kaleidoscope has a narrow scope as it focuses only on exam-based questions and does not evaluate real-world understanding and practical reasoning. Multimodal benchmarks released along with their corresponding models include PangeaBench \cite{yue2025pangea} (14 datasets in 47 languages), and AyaVisionBench \cite{2025ayavision} (9 tasks in 23 languages). However, these sets commonly have few manually verified samples per language, making the results less reliable for a particular target language. \Cref{tab:multilingual_vqa_compact} provides an overview and compares them with our benchmark. A significant portion of our images per language (>1K) and our QA (>1K) are using culturally sourced, which is 5 to 20x larger than the culturally relevant subsets of previous datasets in this field.

\subsection{Indic and Domain‐Specific Datasets}
Although there are dozens of English VQA resources (VQAv2 \cite{balanced_vqa_v2}, CLEVR‐Math \cite{lindstrom2022-clevr-math}, RealWorldQA \cite{xaiorg2024realworldqa}), there is a need for benchmarks in Indian languages. Previous works include \cite{singh-etal-2024-indicgenbench, arora-etal-2023-llms-jeebench}. IndicGenBench \cite{singh-etal-2024-indicgenbench} consists of 5 diverse tasks in 29 Indic languages but does not include multimodality. JEEBench \cite{arora-etal-2023-llms-jeebench} consists of questions from the highly competitive IIT JEE-Advanced exam, but does not include images-related questions. In contrast, our \jee set only includes questions with images. Our work fills the need for an accurate and diverse multimodal benchmark by providing $\sim$4k Visual Question Answering each in Hindi and Telugu.

\begin{figure*}[ht]
    \centering
    \includegraphics[width=\textwidth]{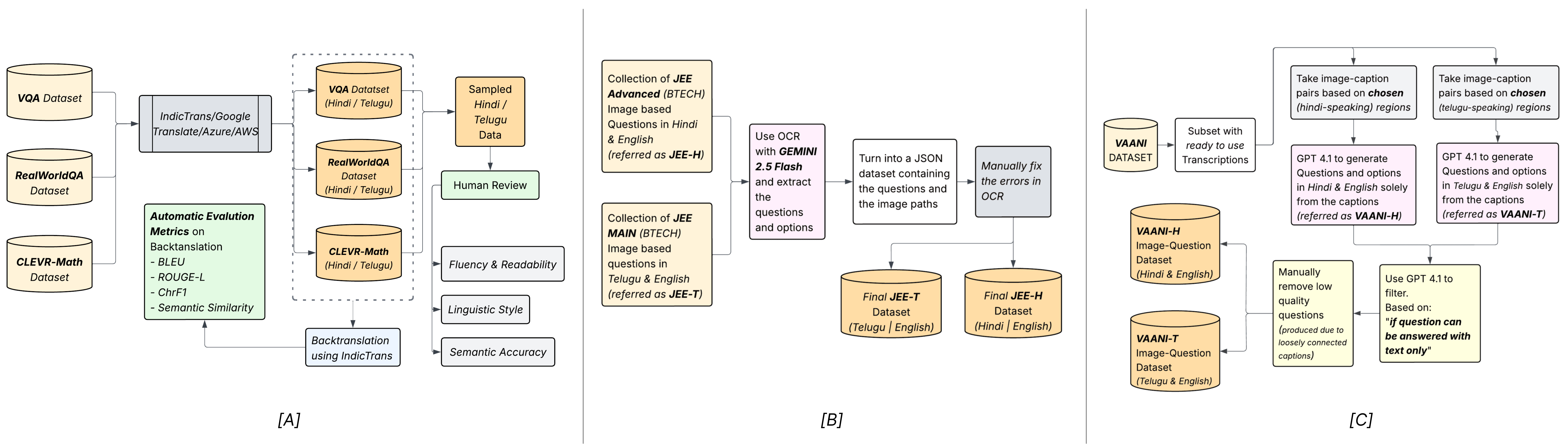} 
    \caption{Dataset generation pipeline for \textbf{[A]} VQAv2, RealWorldQA \& CLEVR-Math \textbf{[B]} VAANI-H \& VAANI-T \textbf{[C]} JEE-H \& JEE-T. We use back translation and text only LLMs to reduce human involvement in QA generation.}
    \label{fig:pipeline}
\end{figure*}

\section{Datasets}
\vspace{-5px}
\subsection{Sources}

\begin{figure*}[ht]
    \centering
    \includegraphics[width=\textwidth]{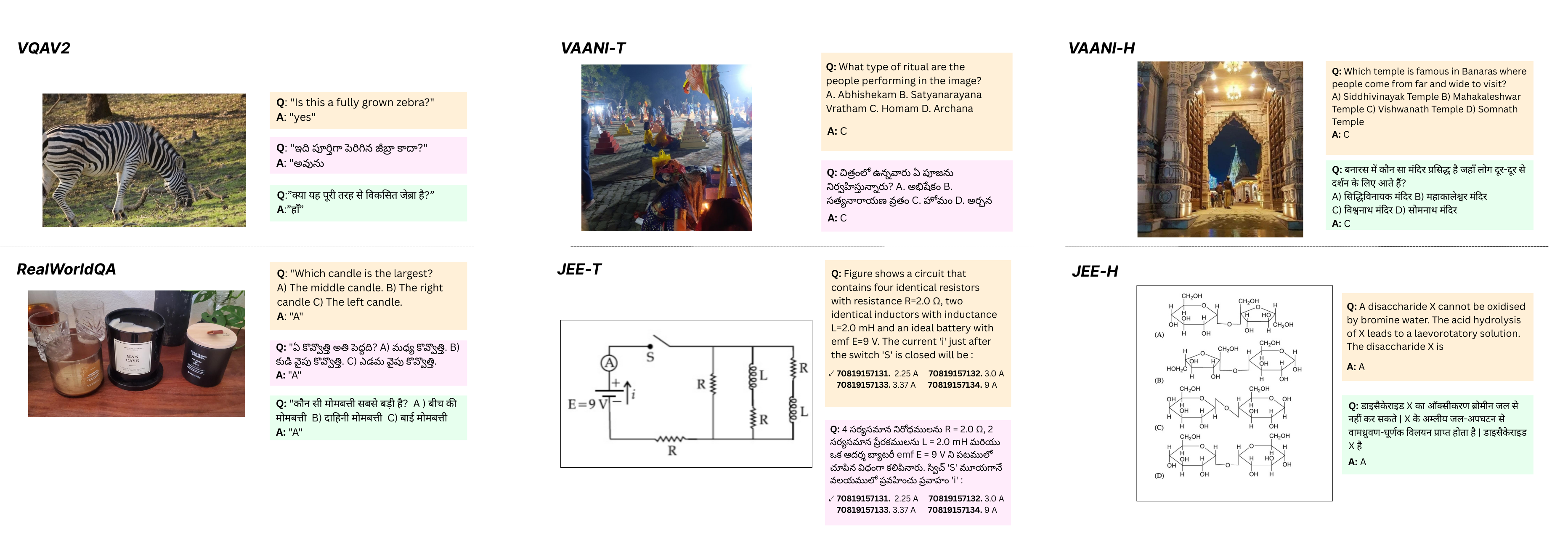} 
    \caption{Qualitative Examples for different domains in our dataset. More images are shown in the appendix}
    \label{fig:allds}
\end{figure*}

Our goal is to create a diverse and robust benchmark for Hindi and Telugu by combining samples translated from English VQA datasets with two new India‑centric datasets, \Cref{fig:allds}. Concretely, we cover the following five broad tasks. (1) \textbf{Real-World Understanding:} We sample 1000 examples from the VQAv2 dataset \cite{Goyal_2017_CVPR} and translate it. VQAv2 is a multimodal dataset designed for visual question answering. (2) \textbf{Practical Visual Reasoning: } We extend the \realworldqa dataset (765 QA) using the translation pipeline. RealWorldQA \cite{xai2024realworldqa,hfrealworldqa} is a benchmark designed to evaluate real-world spatial understanding. (3) \textbf{Visual Mathematical Reasoning: } We sample 1000 samples from the CLEVR-Math dataset and translate it. \clevrmath~\cite{lindstrom2022-clevr-math} is a multimodal dataset designed for basic arithmetic and spatial reasoning. (4) \textbf{STEM Competency testing (JEE-Vision): } The Joint Entrance Exam (JEE) is a high-school nation wide exam conducted yearly in multiple languages, and the exam papers are distributed for the general public (JEE \href{https://jeeadv.ac.in/archive.html}{Adv} and \href{https://www.nta.ac.in/Downloads}{Mains}).
(5) \textbf{Cultural and Region-specific QA: } From the VAANI~\cite{vaani2025} image-caption corpus, we select images sourced from Hindi and Telugu‑speaking regions and use GPT‑4.1 to generate multiple‐choice questions in each language from the original captions to form the VAANI-H and VAANI-T datasets respectively.
We share the prompt to generate this QA in the appendix.


We include existing datasets by translating English sets as opposed to creating completely new datasets from scratch based on the motivations in \cite{singh-etal-2024-indicgenbench}. Translation-based extension of existing benchmark results in multi-way parallel data, allowing researchers to attribute performance due to task knowledge vs. language understanding, and measure cross-lingual generalization. Using the translation pipeline, we are able to leverage the quality control that went into designing the initial benchmarks.

\begin{table}[ht]
\centering
\small
\adjustbox{width=\columnwidth}{
\begin{tabular}{lrrrrrrr}
\toprule
\textbf{Language} & \textbf{VQAv2} & \textbf{RealWorldQA} & \textbf{\clevrmath} & \textbf{JEE-H} & \textbf{JEE-T} & \textbf{VAANI-H} & \textbf{VAANI-T} \\
\midrule
Hindi   & 1,000 & 765 & 1,000 & 192 & -   & 945  & - \\
Telugu  & 1,000 & 765 & 1,000 & -   & 325 & -    & 1,020 \\
English & 1,000 & 765 & 1,000 & 192 & 325 & 945  & 1,020 \\
\midrule
\textbf{Total} & 3,000 & 2,295 & 3,000 & 384 & 650 & 1,890 & 2,040 \\
\bottomrule
\end{tabular}}
\caption{Number of QA pairs per task per language in \datasetname. The samples used in \vqasource, \realworldqa and \clevrmath~ are the same across all languages. For the VAANI and \jee sets, the samples are aligned across the Indian language and English.}
\label{tab:data_languages_sources}
\end{table}

The distribution of the number of QA pairs per language per subset is reported in \Cref{tab:data_languages_sources}.
\subsection{Dataset Generation and translation}
\label{sec:dataset_gen_translation}

\subsection*{Translation Pipeline}
\Cref{fig:pipeline} [A] illustrates our pipeline for translating \vqasource, \realworldqa and \clevrmath~into Hindi and Telugu. 
For each language, we review the translation with 50 diverse samples using  IndicTrans~\cite{gala2023indictrans}, Google Translate, Azure and AWS. 
Based on the performance, we selected the best translation tool per language. 
For Telugu, we used AWS Translate
and for Hindi we used Azure. 
To ensure an accurate evaluation benchmark, we manually review each sample. During manual review, the annotator verifies and makes minor edits based on \textit{Semantic Accuracy}, \textit{Linguistic Style}, and \textit{Fluency \& Readability}, as performed in \cite{kj2025indicmmlu}.
Previous works \cite{kj2025indicmmlu} have used back-translation to the source language in combination with automated evaluation metrics \cite{papineni-etal-2002-bleu, popovic-2015-chrf} to verify a portion of the samples or to select samples to verify. However, we found that the subsets selected using back-translation were biased towards high confidence mistakes of the translation system, thus biasing the dataset distribution.

Using the translation pipeline to generate the initial sample reduces the annotator burden per sample from generating a QA pair to only correcting a translated QA pair. Additionally, it leads to a consistent formatting of the translations. In a representative subset such as VQAv2, 79\% of the samples that were reviewed were accepted during manual review without any change. Within the subset that was changed, about 42\% are minor changes (such as the tense of a word) and only the remaining 58\% need deletion and/or addition of new words during the manual verification stage. Based on our early evaluation on a subset, processing all samples, except those that need addition and deletion, is 5$\times$ faster than generating samples from scratch and equally accurate. Finally, since our translation pipeline reduces the burden and in order to maximize the size of the dataset given a fixed annotator budget, we use 1 annotator per sample. All verifications were carried out by co-authors who are native speakers or possess fluent proficiency in Hindi and Telugu, with sufficient linguistic and contextual understanding to perform high-quality annotations.

\subsection*{VAANI-H and VAANI-T Dataset Generation}
\Cref{fig:pipeline} [C] details the creation process for our VAANI-H and VAANI-T image-question datasets. From the original VAANI set, we extract a subset which have text transcriptions. There are no overlapping set that has both Hindi and Telugu transcriptions and hence we create separate sets of both languages.
For each of the languages, we use text only GPT 4.1 to generate the questions and options in both the languages from their respective captions, resulting in the VAANI-T and VAANI-H for Telugu and Hindi Image-Question datasets respectively. Both datasets then undergo a refinement process: First, text-only GPT 4.1 is used to filter out the question answer pairs that can be answered using only the text without the requirement of the accompanying image. Second, we employ a human verification process to filter out low-quality questions. These were necessary because some of the captions in the VAANI set are relevant to the images and will not be needed for a dataset in which the captions always align with the image.

\subsection*{JEE-H and JEE-T Dataset Creation}
We present JEE-Vision, a novel vision-language dataset constructed from India's Joint Entrance Examination (JEE)\footnote{JEE Advanced:\url{https://jeeadv.ac.in/archive.html}}\footnote{JEE Mains:\url{https://www.nta.ac.in/Downloads}}, leveraging its unique linguistic distribution (JEE Mains: 13 languages; Advanced: English/Hindi) to source authentic STEM problems authored in the target languages by subject-matter experts. This helps us avoiding any manual steps needed to filter out the translation artifacts that compromise existing benchmarks. Unlike text-only evaluations such as \cite{arora-etal-2023-llms-jeebench}, we specifically curate diagram-dependent problems (circuit schematics, chemical structures etc.) creating the first benchmark to evaluate: (1) non-translated multilingual STEM reasoning, and (2) joint understanding of technical visuals and linguistic content in native educational contexts. \Cref{fig:pipeline} [B] outlines the procedure the JEE datasets.
The JEE-H dataset consists of questions from \textit{JEE-Advanced-B.TECH} examination. The dataset has 192 questions (with accompanying images) spanning Math, Physics, Chemistry in both English and Hindi. However, the released questions for the Advanced set are only in English and Hindi. Hence, for the JEE-T dataset, we collect questions from \textit{JEE-Main-B.TECH} examination. The dataset has 325 questions that have accompanying image across Math, Physics, Chemistry in both English and Telugu. For both JEE sets, we extract the questions and options using OCR with Gemini-2.5-Flash~\cite{google2025gemini25}. During initial dataset generation, we found some OCR issues in the samples. Hence, all the questions undergo a manual verification to fix any OCR errors. While this set currently requires a manual verification step, as OCR capabilities improve, this manual verification process will become unnecessary.

\section{Experimental Setup}
\subsection{Selected Models}
We evaluate the performance of current open-weight and frontier proprietary Multimodal Large Language Models (MLLMs) on our benchmark. Some previous works report multilingual results in a target language using models that may not be trained to support that specific language~\cite{2025ayavision, palowacv}. In contrast, we include model evaluations for a target Indian language only if that model claims prior proficiency in the target language. This is to capture an accurate representation of the current multilingual multimodal models.

For Hindi, we report results on open-weight models including Google’s Gemma3 series (Gemma3-4B, 12B, and 27B)~\cite{team2025gemma}, Chitrarth-8B~\cite{khan2024chitrarth}, an Indic-focused vision-language model designed for multiple regional languages including both Hindi and Telugu; Aya-8B~\cite{2025ayavision}, Alibaba's Qwen2.5VL-7B~\cite{bai2025qwen25vl}; and Meta’s LLaMA 3.2 Vision 11B~\cite{meta2025llama32vision}. These models were chosen for their class leading capabilities and explicit support for Hindi. From the set of proprietary models, we report on the Gemini-2.5 Flash variants~\cite{google2025gemini25}, which exhibits strong support for Telugu and Hindi, with notable performance in OCR, and OpenAI’s GPT-4.1~\cite{openai2025gpt41}, a state-of-the-art model with documented support for both Hindi and Telugu.

For Telugu, we find that very few VLMs support the language. We report on the Gemini series (Gemini 1.5 Flash, 2.0 Flash, and 2.5 Flash variants)~\cite{google2025gemini25}, OpenAI’s GPT-4.1~\cite{openai2025gpt41} and Chitrarth-8B~\cite{khan2024chitrarth}, all of which support Telugu.

\renewcommand{\arraystretch}{1.2}
\begin{table*}[t]
\centering
\small
\begin{tabular}{lcc|cc|cc|cc|cc|cc}
\toprule
\textbf{Model} &
\multicolumn{2}{c|}{\textbf{VQAv2}} &
\multicolumn{2}{c|}{\textbf{RealWorldQA}} &
\multicolumn{2}{c|}{\textbf{CLEVR-Math}} &
\multicolumn{2}{c|}{\textbf{JEE-T}} &
\multicolumn{2}{c|}{\textbf{VAANI-T}} &
\multicolumn{2}{c}{\textbf{Ours-T}} \\
\cmidrule(lr){2-3}
\cmidrule(lr){4-5}
\cmidrule(lr){6-7}
\cmidrule(lr){8-9}
\cmidrule(lr){10-11}
\cmidrule(lr){12-13}
& \textbf{Tel} & \textbf{En} & \textbf{Tel} & \textbf{En} & \textbf{Tel} & \textbf{En} & \textbf{Tel} & \textbf{En} & \textbf{Tel} & \textbf{En} & \textbf{Tel} & \textbf{En} \\
\midrule
GPT 4.1 & 68.70 & 72.00 & 61.05 & \textbf{75.29} & 46.60 & \underline{65.00} & 34.36 & 40.92 & \underline{81.57} & \textbf{82.45} & 58.46 & 67.13 \\
Gemini 2.5 Flash & \underline{75.10} & 74.10 & \textbf{61.18} & \underline{73.20} & \textbf{66.70} & \textbf{71.80} & \textbf{45.90} & \textbf{54.15} & \textbf{82.75} & 80.78 & \textbf{66.33} & \textbf{70.81} \\
Gem 2.0 Flash & 70.20 & 74.20 & \underline{60.92} & 69.67 & 43.60 & 53.50 & \underline{42.15} & \underline{53.23} & 80.49 & 79.61 & 59.47 & 66.04 \\
Gem 1.5 Flash & 68.50 & \underline{74.40} & 60.00 & 67.19 & 37.40 & 46.70 & 29.85 & 39.38 & 76.27 & 79.61 & 54.40 & 61.46 \\
Chitrarth & \textbf{76.00} & \textbf{78.50} & 53.59 & 52.55 & \underline{53.90} & 56.90 & 20.00 & 18.15 & \underline{81.57} & \underline{82.45} & 57.01 & 57.71 \\
\midrule
Model Mean & 71.10 & 74.64 & 59.35 & 67.58 & 49.64 & 58.78 & 34.85 & 41.17 & 80.53 & 80.98 & 59.13 & 64.63 \\
\bottomrule
\end{tabular}
\caption{Results (in \%) for Telugu (Tel) and English (En). \textbf{Bold} indicates the best and \underline{underline} indicates the second best. There is a performance regression for all sets for most models from English to Telugu, with the primary exception being VAANI-T. Ours-T includes the VAANI-T, JEE-T subsets along with \vqasource, \clevrmath~and \realworldqa}
\label{tab:multilingual_results_eng_tel}
\end{table*}

\renewcommand{\arraystretch}{1.2}
\begin{table*}[t]
\centering
\small 
\begin{tabular}{lcc|cc|cc|cc|cc|cc}
\toprule
\textbf{Model} &
\multicolumn{2}{c|}{\textbf{VQAv2}} &
\multicolumn{2}{c|}{\textbf{RealWorldQA}} &
\multicolumn{2}{c|}{\textbf{CLEVR-Math}} &
\multicolumn{2}{c|}{\textbf{JEE-H}} &
\multicolumn{2}{c|}{\textbf{VAANI-H}} &
\multicolumn{2}{c}{\textbf{Ours-H}}\\
\cmidrule(lr){2-3}
\cmidrule(lr){4-5}
\cmidrule(lr){6-7}
\cmidrule(lr){8-9}
\cmidrule(lr){10-11}
\cmidrule(lr){12-13}
& \textbf{Hi} & \textbf{En} & \textbf{Hi} & \textbf{En} & \textbf{Hi} & \textbf{En} & \textbf{Hi} & \textbf{En} & \textbf{Hi} & \textbf{En} & \textbf{Hi} & \textbf{En} \\
\midrule
GPT-4.1 & \textbf{68.00} & 72.00 & \textbf{70.59} & \textbf{75.29} & 48.10 & 65.00 & \underline{23.18} & \underline{23.05} & \underline{93.33} & \underline{86.88} & 60.64 & 64.44 \\
Gemini 2.5 Flash & 65.00 & 74.10 & \underline{69.54} & \underline{73.20} & \textbf{60.70} & \underline{71.80} & \textbf{56.90} & \textbf{62.89} & \textbf{93.86} & \textbf{87.19} & \textbf{69.20} & \textbf{73.84} \\
Chitrarth & \underline{66.00} & \textbf{78.50} & 52.94 & 52.55 & \underline{57.20} & 56.90 & 11.72 & 13.93 & 84.23 & 80.14 & 54.42 & 56.40 \\
Qwen2.5VL-7B & 37.20 & \underline{74.30} & 51.11 & 68.10 & 29.10 & \textbf{98.80} & 17.84 & 20.70 & 81.79 & 84.76 & 43.41 & 69.33 \\
Aya-8B & 36.30 & 47.30 & 55.42 & 58.82 & 46.20 & 61.40 & 9.63 & 16.02 & 82.22 & 80.42 & 45.95 & 52.79 \\
LLaMA 3.2 11B & 35.90 & 59.80 & 35.68 & 61.57 & 18.90 & 35.60 & 14.32 & 14.45 & 77.67 & 83.28 & 36.49 & 50.94 \\
Gemma3-27B & 64.10 & 65.50 & 54.38 & 61.04 & 43.50 & 53.70 & 19.66 & 17.58 & 87.41 & 82.01 & 53.81 & 55.97 \\
Gemma3-12B & 63.00 & 65.70 & 53.98 & 58.69 & 40.20 & 46.80 & 14.84 & 16.80 & 85.50 & 82.33 & 51.50 & 54.87 \\
Gemma3-4B & 55.00 & 58.20 & 43.27 & 50.19 & 33.20 & 39.60 & 14.32 & 17.19 & 80.64 & 77.78 & 45.29 & 48.59 \\
\midrule
Model Mean &
53.74 & 66.26 &
53.99 & 62.49 &
43.01 & 58.62 &
20.01 & 22.29 &
85.85 & 83.76 &
51.19 & 58.49 \\
\bottomrule
\end{tabular}
\caption{Results (in \%) for Hindi (Hi) and English (En). \textbf{Bold} indicates the best and \underline{underline} indicates the second best. There is a performance regression for all sets for most models from English to Hindi, with the primary exception being VAANI-H. Ours-H includes the VAANI-H, JEE-H subsets along with \vqasource, \clevrmath~and \realworldqa}
\label{tab:multilingual_results_hindi}
\end{table*}

\subsection{Evaluation Metrics}
\label{section:eval_metrics}

The RealWorldQA, VAANI-H/T subsets have only multiple-choice questions and JEE-T has multiple-choice or integer-answer questions. We use accuracy as the evaluation metric for all of these sets. We extract the answers using regex-based parsing, and report the overall accuracy across all the questions.

\textbf{Hybrid Evaluation for VQA and CLEVR-Math: } For VQAv2 and CLEVR-Math subsets, the answers are either a single word or short phrases. We adopt a hybrid evaluation strategy. We first evaluate a sample using exact match. Our exact match evaluation is built using the official VQA evaluation script~\cite{balanced_vqa_v2}, with the functionalities also extended to Hindi and Telugu. While exact match is strict and interpretable, it may penalize correct answers with minor surface-level variations (e.g., “yes” and “yes, it is”, synonyms, etc.). If exact match fails for a sample, we evaluate that sample using "gpt-4.1-2025-04-14" \cite{openai2025gpt41} as descr in \cite{Judging_LLM-as-a-judge}. This two-step approach enables both high precision and flexibility, especially in cases in which answers may vary in form but not meaning. Its impact on scores is discussed in the appendix.

\textbf{JEE-H Evaluation: } The JEE-H Dataset has single correct MCQs, multiple correct MCQs, and  numeric-type questions. We extend the scoring process used in ~\cite{arora-etal-2023-llms-jeebench}. However, instead of the manual step they use, we replace with regex-based parsing to extract answers, and rule based processing to score each question. See appendix for details.

\begin{table*}[t]
\centering
\small 
\begin{tabular}{lcc|cc|cc|cc|cc|cc}
\toprule
\textbf{Method} &
\multicolumn{2}{c|}{\textbf{VQAv2}} &
\multicolumn{2}{c|}{\textbf{RealWorldQA}} &
\multicolumn{2}{c|}{\textbf{JEE-H}} &
\multicolumn{2}{c|}{\textbf{CLEVR-Math}} &
\multicolumn{2}{c}{\textbf{VAANI-H}} &
\multicolumn{2}{c}{\textbf{Ours-H}} \\
\cmidrule(lr){2-3}
\cmidrule(lr){4-5}
\cmidrule(lr){6-7}
\cmidrule(lr){8-9}
\cmidrule(lr){10-11}
\cmidrule(lr){12-13}
& \textbf{Hi} & \textbf{En} & \textbf{Hi} & \textbf{En} & \textbf{Hi} & \textbf{En} & \textbf{Hi} & \textbf{En} & \textbf{Hi} & \textbf{En} & \textbf{Hi} & \textbf{En} \\
\midrule
Standard & \textbf{64.10} & \textbf{65.50} & \underline{54.38} & \underline{61.04} & \underline{19.66} & \underline{17.58} & \underline{43.50} & \underline{53.70} & \textbf{87.41} & \underline{82.01} & 53.81 & 55.97 \\
CoT & \underline{60.50} & 62.70 & \textbf{61.96} & \textbf{64.31} & \textbf{21.88} & \textbf{31.38} & \textbf{44.3} & \textbf{57.1} & \underline{84.34} & \textbf{83.28} & \textbf{54.60}&\textbf{59.75} \\
Caption-Only & 56.40 & \underline{64.1} & 44.05 & 54.77 & -- & -- & 36.4 & 32.9 & 80.85 & 77.14 & 
-- & --
\\
\bottomrule
\end{tabular}
\caption{Results on Gemma-27B across multiple datasets using difference inference methods. \textbf{Bold} indicates the best performance, \underline{underline} indicates the second best. CoT significantly improves results on English but has a much smaller improvement when using Hindi.}
\label{tab:gemma27b_ablation}
\end{table*}

\section{Results and Analysis}

\subsection{English-Telugu Comparison}
The results of our evaluated models for Telugu and English are reported in \Cref{tab:multilingual_results_eng_tel}. In all the tasks, the average performance of the models in English is higher than in Telugu, though the margin on the VAANI-T subset is marginal. Averaging across the models and the tasks, there is a gap of 5.5 points between the English and Telugu subsets. 

Results on VAANI-T go against the trend of the other subsets and we explore this in \Cref{sec:task-specific}. Overall, the best performing model is Gemini 2.5 Flash~\cite{google2025gemini25} on both Telugu and English. A notable model was Chitrarth~\cite{khan2024chitrarth}, which has been trained on \vqasource~in multiple languages and hence the leader on the VQAv2 subset in English and Telugu. GPT-4.1 shows leading on the \realworldqa~subset for English and Hindi but much poorer performance on the Telugu version. This data point demonstrates how models can have a high performance in some languages but fail to generalize and sheds light on the need for thorough evaluation of all the target languages.

\subsection{English-Hindi Comparison}
\label{sec:eng-hin-results}
The results of the evaluated models for Hindi and English are in \Cref{tab:multilingual_results_hindi}. For 4 out of 5 tasks, the average performance in English is higher than in Hindi. This demonstrates the need to improve the capabilities of our multilingual LLMs. Averaging across the models and the tasks, there is a gap of 8.3 points between the English and Hindi subsets.

Going against the trend, results on VAANI-H are higher in Hindi than in English, further discussed in \Cref{sec:task-specific}. Among the proprietary and open-weight models, the open source models lag the proprietary ones in both Hindi and English in most cases. Gemini 2.5 Flash~\cite{google2025gemini25} is the overall best performing model on both languages in this paired dataset. Among the open-weight models, the largest gap is seen on Qwen2.5VL-7B~\cite{bai2025qwen25vl}, dropping from 25.92 points from English and Hindi.

\subsection{Telugu-Hindi Comparison}

We have aligned samples on the \vqasource, \clevrmath and \realworldqa sets for Hindi, Telugu and English along with evaluations with GPT-4.1, Gemini 2.5 Flash and Chitrarth for all three languages. On this subset of datasets and models, the average performance on Hindi, Telugu and English is 61.51, 62.53 and 68.6 respectively.
This comparison demonstrates that the performance regresses similarly from English to both Indian languages and underlines a systemic issues when using the current frontier model with Indian languages.

\subsection{Task-Specific Results}
\label{sec:task-specific}
\begin{figure}[ht]
    \centering
    \includegraphics[width=0.95\columnwidth]{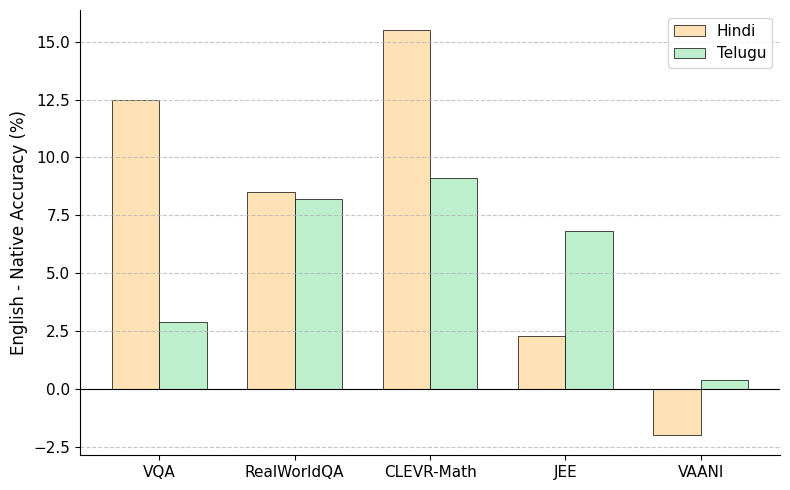}
    \caption{Average performance regression from English to Indic Language per domain across all models. In all but one scenario, there is a regression in performance from English to the target Indian language}
    \label{fig:per-language-per-dataset-regression}
\end{figure}

The average gaps across models for each of the tasks is visualized in \Cref{fig:per-language-per-dataset-regression}. This graph motivates the need for overall, multi-dimensional improvement needed in the capabilities of multilingual LLMs to catch up to the performance in English. \clevrmath~and \realworldqa~have the largest deltas in performance between the English and the Indic Language sets.
In comparison, the smallest delta in performance is on the VAANI dataset. On VAANI-T, average performance is marginally worse than English but on VAANI-H, the average performance is 2.09 points higher on the Hindi subset. On reviewing the QA that have the correct prediction for Indic Languages and incorrect predictions for English, we found two possible reasons.  Firstly, we found samples where the English QA generated did not completely capture the meaning of the options or translated the options into English for OCR tasks where the text in the Indic language was visible in the image. This pattern made the questions harder for the model in English sets.
Secondly, we find that text-only LLMs are poor at generating distractors when generating MCQA from VAANI captions. Thus, the models may guess the correct answer by exploiting statistical patterns, which inflates metrics. In future work, we'll explore using LLM based quality control to further align questions across languages and using Multi-Binary Accuracy~\cite{cai2024temporalbench} for VAANI subsets.

\subsection{Inference Techniques and Baselines}
We ablate prompting and other baselines using the Gemma-27B model on the Hindi subsets. We report comparisons for Direct Inference (Standard Prompting), Chain-of-Thought \cite{wei2022chain}, and a Caption-Only \cite{deitke2024molmo} baseline, \Cref{tab:gemma27b_ablation}.

Standard prompting provides a strong baseline performance across all datasets, especially on VAANI and VQA, indicating that the model benefits significantly from both modalities in relatively straightforward recognition tasks. In contrast, Chain-of-Thought (CoT) prompting notably improves performance on reasoning-heavy datasets. On JEE-H, CoT boosts performance in English and Hindi by 13.8 and 2.22 points respectively. Similar gains are observed in \clevrmath, where CoT surpasses the standard setup in both languages. These improvements highlight the effectiveness of step-by-step reasoning prompts in tasks that require deeper cognitive processing. However, CoT does not benefit all tasks, with marginal gains or regression on VAANI and \vqasource, where reasoning depth is less critical. CoT may be introducing unnecessary verbosity or distraction when the task primarily relies on perception. Overall, CoT benefits the English subset by 3.78 points in comparison to the benefit on Hindi at 0.79 points. This indicates a possible bias in the training data, with possibly more training done on English CoT data.

Next, to explore the need for visual embedding for QA, we evaluate a caption-only baseline. Here, the Gemma-27B model is asked to first generate a generate a rich textual summary of the image. Then the generated caption and the original question are given to the model without the original image. This approach is inferior to the direct image-based inputs in all tasks.

\subsection{Failure Analysis of Visual Language Models}

We categorized the failures of GPT 4.1 on VAANI-T subset for both English and Telugu. We analyze results on VAANI-T as it sources images and QA from the Indian context and failures on this set point to common reasons for errors for current multimodal multilingual LLMs. We found 4 major categories of errors that appeared in both English and Telugu questions. They are: (1) Missing Indian context: Failed predictions in this category were due to missing knowledge specific to India needed to answer the question. (2) Visual Grounding error: Failures in this category were due to the model's inability to associate objects in the picture with the question or options. (3) Visual Perception Failure: Errors in this category are due to the model failing to interpret the content of the image correctly. (4) Failure to Ground in Indian Context: Samples in this category accurately identified visual elements but they failed to analyze them in the appropriate socio-cultural context. \Cref{tab:error_categories_by_language} reports the percentages of the different categories on this dataset.


\begin{table}[ht]
\centering
\small
\adjustbox{width=\columnwidth}{
\begin{tabular}{lrr}
\toprule
\textbf{Error Category} & \textbf{English} & \textbf{Telugu} \\
\midrule
Lack of Knowledge about India          & 17\% & 16\% \\
Visual Grounding Error & 49\% & 50\% \\
Visual Perception Failure           & 19\% & 18\% \\
Failure to Ground in Indian Culture           & 15\% & 16\% \\
\bottomrule
\end{tabular}}
\caption{Error category distribution for GPT 4.1 on VAANI-T for English and Telugu. Percentages indicate the proportion of errors falling into each category.}
\label{tab:error_categories_by_language}
\end{table}

\subsection{Discussion and Future Work}
Using our framework, benchmarks can be created in a number of domains. We aim to extend the framework as follows. First, the proprietary models achieve high scores on the VAANI-H/T. We hypothesize that the text only LLMs we use for generating QA from captions do not design good distractors. Thus, the VLMs may guess the correct answer by exploiting statistical patterns. A future work is using Multi-Binary Accuracy~\cite{cai2024temporalbench} for VAANI subsets. Second, the JEE-Advanced examination data source only contains 2 image-based Mathematics questions and we plan to include the questions in JEE-Mains. 
Our current scope is limited to images but can be extended to video understanding similar to \cite{shafique2025culturallydiversemultilingualmultimodalvideo}. Finally, there are 122 major Indian languages, and we cover only Hindi and Telugu. We hope this benchmark gets extended to all other Indic languages with contributions from native speakers from those languages. 

\section{Conclusion}
This paper introduces \datasetname, a framework for developing benchmarks to evaluate multimodal large language models in Hindi and Telugu, addressing critical gaps in existing multilingual evaluations. We combined semi-automated dataset creation with rigorous human verification and sourced culturally grounded native datasets to assess diverse capabilities. Evaluations of state-of-the-art VLMs reveal significant performance regressions in Indic languages compared to English, emphasizing the need for targeted improvements in multilingual visual understanding. The regression highlights the need for rigorous evaluation for all Indic languages, beyond Hindi and Telugu.


\bibliography{custom}

@misc{ahuja2023megamultilingualevaluationgenerative,
      title={MEGA: Multilingual Evaluation of Generative AI}, 
      author={Kabir Ahuja and Harshita Diddee and Rishav Hada and Millicent Ochieng and Krithika Ramesh and Prachi Jain and Akshay Nambi and Tanuja Ganu and Sameer Segal and Maxamed Axmed and Kalika Bali and Sunayana Sitaram},
      year={2023},
      eprint={2303.12528},
      archivePrefix={arXiv},
      primaryClass={cs.CL},
      url={https://arxiv.org/abs/2303.12528}, 
}

@article{2025ayavision,
  title={Aya Vision: Advancing the Frontier of Multilingual Multimodality},
  author={Saurabh Dash and Yiyang Nan and John Dang and Arash Ahmadian and Shivalika Singh and Madeline Smith and Bharat Venkitesh and Vlad Shmyhlo and Viraat Aryabumi and Walter Beller-Morales and Jeremy Pekmez and Jason Ozuzu and Pierre Richemond and Acyr Locatelli and Nick Frosst and Phil Blunsom and Aidan Gomez and Ivan Zhang and Marzieh Fadaee and Manoj Govindassamy and Sudip Roy and Matthias Gallé and Beyza Ermis and Ahmet Üstün and Sara Hooker},
  journal={arXiv preprint arXiv:2505.08751},
  volume = {arXiv:2505.08751},
  url = {http://arxiv.org/abs/2505.08751},
  year={2025}
}

@misc{vaani2025,
  author       = {VAANI Team},
  title        = {VAANI: Capturing the Language Landscape for an Inclusive Digital India (Phase 1)},
  url = {https://vaani.iisc.ac.in/},
  howpublished = {\url{https://vaani.iisc.ac.in/}},
  year         = {2025}
}

@inproceedings{
    yue2025pangea,
    title={Pangea: A Fully Open Multilingual Multimodal {LLM} for 39 Languages},
    author={Xiang Yue and Yueqi Song and Akari Asai and Seungone Kim and Jean de Dieu Nyandwi and Simran Khanuja and Anjali Kantharuban and Lintang Sutawika and Sathyanarayanan Ramamoorthy and Graham Neubig},
    booktitle={The Thirteenth International Conference on Learning Representations},
    year={2025},
    url={https://openreview.net/forum?id=a3g2l4yEys}
}

@article{wu2025bitterlessonlearned2000,
      title={The Bitter Lesson Learned from 2,000+ Multilingual Benchmarks}, 
      author={Minghao Wu and Weixuan Wang and Sinuo Liu and Huifeng Yin and Xintong Wang and Yu Zhao and Chenyang Lyu and Longyue Wang and Weihua Luo and Kaifu Zhang},
      journal={arXiv preprint arXiv:2504.15521},
      volume={arXiv:2504.15521},
      url={https://arxiv.org/abs/2504.15521}, 
      year={2025},
}

@inproceedings{palowacv,
  author={Rasheed, Hanoona and Maaz, Muhammad and Shaker, Abdelrahman and Khan, Salman and Cholakal, Hisham and Anwer, Rao M. and Baldwin, Tim and Felsberg, Michael and Khan, Fahad S.},
  booktitle={Winter Conference on Applications of Computer Vision (WACV)}, 
  title={Palo: A Polyglot Large Multimodal Model for 5B People}, 
  year={2025},
  pages={1745-1754},
  doi={10.1109/WACV61041.2025.00177}
}

@inproceedings{singh-etal-2024-indicgenbench,
    title = "{I}ndic{G}en{B}ench: A Multilingual Benchmark to Evaluate Generation Capabilities of {LLM}s on {I}ndic Languages",
    author = "Singh, Harman  and
      Gupta, Nitish  and
      Bharadwaj, Shikhar  and
      Tewari, Dinesh  and
      Talukdar, Partha",
    editor = "Ku, Lun-Wei  and
      Martins, Andre  and
      Srikumar, Vivek",
    booktitle = "Proceedings of the 62nd Annual Meeting of the Association for Computational Linguistics (Volume 1: Long Papers)",
    month = aug,
    year = "2024",
    address = "Bangkok, Thailand",
    publisher = "Association for Computational Linguistics",
    url = "https://aclanthology.org/2024.acl-long.595/",
    doi = "10.18653/v1/2024.acl-long.595",
    pages = "11047--11073"
}

@article{llava-liu2023visual,
  title={Visual instruction tuning},
  author={Liu, Haotian and Li, Chunyuan and Wu, Qingyang and Lee, Yong Jae},
  journal={Advances in neural information processing systems},
  volume={36},
  pages={34892--34916},
  year={2023}
}

@inproceedings{
    Dai2023InstructBLIPTG,
    title={Instruct{BLIP}: Towards General-purpose Vision-Language Models with Instruction Tuning},
    author={Wenliang Dai and Junnan Li and Dongxu Li and Anthony Tiong and Junqi Zhao and Weisheng Wang and Boyang Li and Pascale Fung and Steven Hoi},
    booktitle={Thirty-seventh Conference on Neural Information Processing Systems},
    year={2023},
    url={https://openreview.net/forum?id=vvoWPYqZJA}
}

@inproceedings{blip2-li2023blip,
  title={Blip-2: Bootstrapping language-image pre-training with frozen image encoders and large language models},
  author={Li, Junnan and Li, Dongxu and Savarese, Silvio and Hoi, Steven},
  booktitle={International conference on machine learning},
  pages={19730--19742},
  year={2023},
  organization={PMLR}
}

@inproceedings{
    minigpt-4,
    title={Mini{GPT}-4: Enhancing Vision-Language Understanding with Advanced Large Language Models},
    author={Deyao Zhu and Jun Chen and Xiaoqian Shen and Xiang Li and Mohamed Elhoseiny},
    booktitle={The Twelfth International Conference on Learning Representations},
    year={2024},
    url={https://openreview.net/forum?id=1tZbq88f27}
}

@inproceedings{romero2024cvqa,
      title={CVQA: Culturally-diverse Multilingual Visual Question Answering Benchmark}, 
      author={David Romero and Chenyang Lyu and Haryo Akbarianto Wibowo and Teresa Lynn and Injy Hamed and Aditya Nanda Kishore and Aishik Mandal and Alina Dragonetti and Artem Abzaliev and Atnafu Lambebo Tonja and Bontu Fufa Balcha and Chenxi Whitehouse and Christian Salamea and Dan John Velasco and David Ifeoluwa Adelani and David Le Meur and Emilio Villa-Cueva and Fajri Koto and Fauzan Farooqui and Frederico Belcavello and Ganzorig Batnasan and Gisela Vallejo and Grainne Caulfield and Guido Ivetta and Haiyue Song and Henok Biadglign Ademtew and Hernán Maina and Holy Lovenia and Israel Abebe Azime and Jan Christian Blaise Cruz and Jay Gala and Jiahui Geng and Jesus-German Ortiz-Barajas and Jinheon Baek and Jocelyn Dunstan and Laura Alonso Alemany and Kumaranage Ravindu Yasas Nagasinghe and Luciana Benotti and Luis Fernando D'Haro and Marcelo Viridiano and Marcos Estecha-Garitagoitia and Maria Camila Buitrago Cabrera and Mario Rodríguez-Cantelar and Mélanie Jouitteau and Mihail Mihaylov and Mohamed Fazli Mohamed Imam and Muhammad Farid Adilazuarda and Munkhjargal Gochoo and Munkh-Erdene Otgonbold and Naome Etori and Olivier Niyomugisha and Paula Mónica Silva and Pranjal Chitale and Raj Dabre and Rendi Chevi and Ruochen Zhang and Ryandito Diandaru and Samuel Cahyawijaya and Santiago Góngora and Soyeong Jeong and Sukannya Purkayastha and Tatsuki Kuribayashi and Thanmay Jayakumar and Tiago Timponi Torrent and Toqeer Ehsan and Vladimir Araujo and Yova Kementchedjhieva and Zara Burzo and Zheng Wei Lim and Zheng Xin Yong and Oana Ignat and Joan Nwatu and Rada Mihalcea and Thamar Solorio and Alham Fikri Aji},
     booktitle = {Advances in Neural Information Processing Systems},
     editor = {A. Globerson and L. Mackey and D. Belgrave and A. Fan and U. Paquet and J. Tomczak and C. Zhang},
     pages = {11479--11505},
     publisher = {Curran Associates, Inc.},
     url = {https://proceedings.neurips.cc/paper_files/paper/2024/file/1568882ba1a50316e87852542523739c-Paper-Datasets_and_Benchmarks_Track.pdf},
     volume = {37},
     year = {2024}
}

@article{tang2024mtvqa,
  title={Mtvqa: Benchmarking multilingual text-centric visual question answering},
  author={Tang, Jingqun and Liu, Qi and Ye, Yongjie and Lu, Jinghui and Wei, Shu and Lin, Chunhui and Li, Wanqing and Mahmood, Mohamad Fitri Faiz Bin and Feng, Hao and Zhao, Zhen and others},
  journal={arXiv preprint arXiv:2405.11985},
  year={2024}
}

@inproceedings{pfeiffer-etal-2022-xgqa,
    title = "x{GQA}: Cross-Lingual Visual Question Answering",
    author = "Pfeiffer, Jonas  and
      Geigle, Gregor  and
      Kamath, Aishwarya  and
      Steitz, Jan-Martin O.  and
      Roth, Stefan  and
      Vuli{\'c}, Ivan  and
      Gurevych, Iryna",
    editor = "Muresan, Smaranda  and
      Nakov, Preslav  and
      Villavicencio, Aline",
    booktitle = "Findings of the Association for Computational Linguistics: ACL 2022",
    month = may,
    year = "2022",
    address = "Dublin, Ireland",
    publisher = "Association for Computational Linguistics",
    url = "https://aclanthology.org/2022.findings-acl.196/",
    doi = "10.18653/v1/2022.findings-acl.196",
    pages = "2497--2511"
}

@article{hudson2018gqa,
    title={GQA: A New Dataset for Real-World Visual Reasoning 
    and Compositional Question Answering},
    author={Hudson, Drew A and Manning, Christopher D},
    journal={Conference on Computer Vision and Pattern Recognition (CVPR)},
    year={2019}
}

@article{
    gala2023indictrans,
    title={IndicTrans2: Towards High-Quality and Accessible Machine Translation Models for all 22 Scheduled Indian Languages},
    author={Jay Gala and Pranjal A Chitale and A K Raghavan and Varun Gumma and Sumanth Doddapaneni and Aswanth Kumar M and Janki Atul Nawale and Anupama Sujatha and Ratish Puduppully and Vivek Raghavan and Pratyush Kumar and Mitesh M Khapra and Raj Dabre and Anoop Kunchukuttan},
    journal={Transactions on Machine Learning Research},
    issn={2835-8856},
    year={2023},
    url={https://openreview.net/forum?id=vfT4YuzAYA},
}

@inproceedings{
    changpinyo2022maxm,
    title={Ma{XM}: Towards Multilingual Visual Question Answering},
    author={Soravit Changpinyo and Linting Xue and Michal Yarom and Ashish V Thapliyal and Idan Szpektor and Julien Amelot and Xi Chen and Radu Soricut},
    booktitle={The 2023 Conference on Empirical Methods in Natural Language Processing},
    year={2023},
    url={https://openreview.net/forum?id=rhGh8jLOPd}
}

@misc{salazar2025kaleidoscopeinlanguageexamsmassively,
      title={Kaleidoscope: In-language Exams for Massively Multilingual Vision Evaluation}, 
      author={Israfel Salazar and Manuel Fernández Burda and Shayekh Bin Islam and Arshia Soltani Moakhar and Shivalika Singh and Fabian Farestam and Angelika Romanou and Danylo Boiko and Dipika Khullar and Mike Zhang and Dominik Krzemiński and Jekaterina Novikova and Luísa Shimabucoro and Joseph Marvin Imperial and Rishabh Maheshwary and Sharad Duwal and Alfonso Amayuelas and Swati Rajwal and Jebish Purbey and Ahmed Ruby and Nicholas Popovič and Marek Suppa and Azmine Toushik Wasi and Ram Mohan Rao Kadiyala and Olga Tsymboi and Maksim Kostritsya and Bardia Soltani Moakhar and Gabriel da Costa Merlin and Otávio Ferracioli Coletti and Maral Jabbari Shiviari and MohammadAmin farahani fard and Silvia Fernandez and María Grandury and Dmitry Abulkhanov and Drishti Sharma and Andre Guarnier De Mitri and Leticia Bossatto Marchezi and Johan Obando-Ceron and Nazar Kohut and Beyza Ermis and Desmond Elliott and Enzo Ferrante and Sara Hooker and Marzieh Fadaee},
      year={2025},
      eprint={2504.07072},
      archivePrefix={arXiv},
      primaryClass={cs.CL},
      url={https://arxiv.org/abs/2504.07072}, 
}

@InProceedings{balanced_vqa_v2,
    author = {Yash Goyal and Tejas Khot and Douglas Summers{-}Stay and Dhruv Batra and Devi Parikh},
    title = {Making the {V} in {VQA} Matter: Elevating the Role of Image Understanding in {V}isual {Q}uestion {A}nswering},
    booktitle = {Conference on Computer Vision and Pattern Recognition (CVPR)},
    year = {2017},
}

@misc{lindstrom2022-clevr-math,
  doi = {10.48550/ARXIV.2208.05358},
  url = {https://arxiv.org/abs/2208.05358},
  author = {Adam Dahlgren Lindstr{\"o}m and Savitha Sam Abraham},
  title = {CLEVR-Math: A Dataset for Compositional Language, Visual, and Mathematical Reasoning},
  publisher = {arXiv},
  year = {2022},
  copyright = {Creative Commons Attribution Share Alike 4.0 International}
}

@misc{xaiorg2024realworldqa,
  title={RealWorldQA: A Benchmark for Real-World Spatial Understanding Capabilities of Multimodal AI Models},
  author={XAI.org},
  year={2024},
  howpublished={\url{https://huggingface.co/datasets/xai-org/RealworldQA}},
  note={RealWorldQA is a benchmark designed for real-world understanding with 765 multiple-choice questions requiring recognition of details in high-resolution images}
}

@article{bai2025qwen25vl,
  title={Qwen2.5-VL Technical Report},
  author={Bai, Shuai and Chen, Keqin and Liu, Xuejing and Wang, Jialin and Ge, Wenbin and Song, Sibo and Dang, Kai and Wang, Peng and Wang, Shijie and Tang, Jun and Zhong, Humen and Zhu, Yuanzhi and Yang, Mingkun and Li, Zhaohai and Wan, Jianqiang and Wang, Pengfei and Ding, Wei and Fu, Zheren and Xu, Yiheng and Ye, Jiabo and Zhang, Xi and Xie, Tianbao and Cheng, Zesen and Zhang, Hang and Yang, Zhibo and Xu, Haiyang and Lin, Junyang},
  journal={arXiv preprint arXiv:2502.13923},
  year={2025},
  url={https://arxiv.org/abs/2502.13923}
}

@article{zhu2025internvl3,
  title={InternVL3: Exploring Advanced Training and Test-Time Recipes for Open-Source Multimodal Models},
  author={Zhu, Jinguo and others},
  journal={arXiv preprint arXiv:2504.10479},
  year={2025},
  url={https://arxiv.org/abs/2504.10479}
}

@misc{meta2025llama32vision,
  title={Llama 3.2-Vision: Instruction-tuned image reasoning generative models},
  author={{Meta Llama}},
  year={2025},
  howpublished={Model release by Meta},
  note={Available at: https://huggingface.co/meta-llama/Llama-3.2-11B-Vision}
}

@article{deitke2024molmo,
  title={Molmo and PixMo: Open Weights and Open Data for State-of-the-Art Vision-Language Models},
  author={Matt Deitke and Christopher Clark and Sangho Lee and Rohun Tripathi and Yue Yang and Jae Sung Park and Mohammadreza Salehi and Niklas Muennighoff and Kyle Lo and Luca Soldaini and Jiasen Lu and Taira Anderson and Erin Bransom and Kiana Ehsani and Huong Ngo and YenSung Chen and Ajay Patel and Mark Yatskar and Chris Callison-Burch and Andrew Head and Rose Hendrix and Favyen Bastani and Eli VanderBilt and Nathan Lambert and Yvonne Chou and Arnavi Chheda and Jenna Sparks and Sam Skjonsberg and Michael Schmitz and Aaron Sarnat and Byron Bischoff and Pete Walsh and Chris Newell and Piper Wolters and Tanmay Gupta and Kuo-Hao Zeng and Jon Borchardt and Dirk Groeneveld and Crystal Nam and Sophie Lebrecht and Caitlin Wittlif and Carissa Schoenick and Oscar Michel and Ranjay Krishna and Luca Weihs and Noah A. Smith and Hannaneh Hajishirzi and Ross Girshick and Ali Farhadi and Aniruddha Kembhavi},
  journal={Conference on Computer Vision and Pattern Recognition (CVPR)},
  year={2025},
  url={https://arxiv.org/abs/2409.17146}
}

@misc{anthropic2025claude37,
  title={Claude 3.7 Sonnet: The First Hybrid Reasoning Model},
  author={{Anthropic}},
  year={2025},
  month={February},
  note={Available at: https://www.anthropic.com/news/claude-3-7-sonnet},
  howpublished={Anthropic Company Website}
}

@misc{openai2025gpt41,
  title={GPT-4.1: A New Series of GPT Models with Major Improvements on Coding, Instruction Following, and Long Context},
  author={{OpenAI}},
  year={2025},
  month={April},
  note={Released on April 14, 2025},
  howpublished={OpenAI Company Website}
}

@inproceedings{papineni-etal-2002-bleu,
    title = "{B}leu: a Method for Automatic Evaluation of Machine Translation",
    author = "Papineni, Kishore  and
      Roukos, Salim  and
      Ward, Todd  and
      Zhu, Wei-Jing",
    editor = "Isabelle, Pierre  and
      Charniak, Eugene  and
      Lin, Dekang",
    booktitle = "Proceedings of the 40th Annual Meeting of the Association for Computational Linguistics",
    month = jul,
    year = "2002",
    address = "Philadelphia, Pennsylvania, USA",
    publisher = "Association for Computational Linguistics",
    url = "https://aclanthology.org/P02-1040/",
    doi = "10.3115/1073083.1073135",
    pages = "311--318"
}

@inproceedings{popovic-2015-chrf,
    title = "chr{F}: character n-gram {F}-score for automatic {MT} evaluation",
    author = "Popovi{\'c}, Maja",
    editor = "Bojar, Ond{\v{r}}ej  and
      Chatterjee, Rajan  and
      Federmann, Christian  and
      Haddow, Barry  and
      Hokamp, Chris  and
      Huck, Matthias  and
      Logacheva, Varvara  and
      Pecina, Pavel",
    booktitle = "Proceedings of the Tenth Workshop on Statistical Machine Translation",
    month = sep,
    year = "2015",
    address = "Lisbon, Portugal",
    publisher = "Association for Computational Linguistics",
    url = "https://aclanthology.org/W15-3049/",
    doi = "10.18653/v1/W15-3049",
    pages = "392--395"
}

@inproceedings{liu-etal-2021-visually-MarVL,
    title = "Visually Grounded Reasoning across Languages and Cultures",
    author = "Liu, Fangyu  and
      Bugliarello, Emanuele  and
      Ponti, Edoardo Maria  and
      Reddy, Siva  and
      Collier, Nigel  and
      Elliott, Desmond",
    editor = "Moens, Marie-Francine  and
      Huang, Xuanjing  and
      Specia, Lucia  and
      Yih, Scott Wen-tau",
    booktitle = "Proceedings of the 2021 Conference on Empirical Methods in Natural Language Processing",
    month = nov,
    year = "2021",
    address = "Online and Punta Cana, Dominican Republic",
    publisher = "Association for Computational Linguistics",
    url = "https://aclanthology.org/2021.emnlp-main.818/",
    doi = "10.18653/v1/2021.emnlp-main.818",
    pages = "10467--10485"
}

@misc{google2025gemini25,
  title={Gemini 2.5: Our most intelligent AI model},
  author={{Google DeepMind}},
  year={2025},
  month={March},
  note={Released on March 25, 2025},
  howpublished={Google DeepMind Blog},
  url={https://blog.google/technology/google-deepmind/gemini-model-thinking-updates-march-2025/}
}

@article{team2025gemma,
  title={Gemma 3 technical report},
  author={Team, Gemma and Kamath, Aishwarya and Ferret, Johan and Pathak, Shreya and Vieillard, Nino and Merhej, Ramona and Perrin, Sarah and Matejovicova, Tatiana and Ram{\'e}, Alexandre and Rivi{\`e}re, Morgane and others},
  journal={arXiv preprint arXiv:2503.19786},
  year={2025}
}

@article{alam2025behind-maya-vlm,
  title={Behind Maya: Building a Multilingual Vision Language Model},
  author={Alam, Nahid and Kanjula, Karthik Reddy and Guthikonda, Surya and Chung, Timothy and Vegesna, Bala Krishna S and Das, Abhipsha and Susevski, Anthony and Chan, Ryan Sze-Yin and Uddin, SM and Islam, Shayekh Bin and others},
  journal={arXiv preprint arXiv:2505.08910},
  year={2025}
}

@inproceedings{arora-etal-2023-llms-jeebench,
    title = "Have {LLM}s Advanced Enough? A Challenging Problem Solving Benchmark For Large Language Models",
    author = "Arora, Daman  and
      Singh, Himanshu  and
      {Mausam}",
    editor = "Bouamor, Houda  and
      Pino, Juan  and
      Bali, Kalika",
    booktitle = "Proceedings of the 2023 Conference on Empirical Methods in Natural Language Processing",
    month = dec,
    year = "2023",
    address = "Singapore",
    publisher = "Association for Computational Linguistics",
    url = "https://aclanthology.org/2023.emnlp-main.468",
    doi = "10.18653/v1/2023.emnlp-main.468",
    pages = "7527--7543",
}

@inproceedings{li2023evaluating,
    title = "Evaluating Object Hallucination in Large Vision-Language Models",
    author = "Li, Yifan  and
      Du, Yifan  and
      Zhou, Kun  and
      Wang, Jinpeng  and
      Zhao, Xin  and
      Wen, Ji-Rong",
    editor = "Bouamor, Houda  and
      Pino, Juan  and
      Bali, Kalika",
    booktitle = "Proceedings of the 2023 Conference on Empirical Methods in Natural Language Processing",
    month = dec,
    year = "2023",
    address = "Singapore",
    publisher = "Association for Computational Linguistics",
    url = "https://aclanthology.org/2023.emnlp-main.20/",
    doi = "10.18653/v1/2023.emnlp-main.20",
    pages = "292--305"
}

@inproceedings{yu2024mmvetevaluatinglargemultimodal,
author = {Yu, Weihao and Yang, Zhengyuan and Li, Linjie and Wang, Jianfeng and Lin, Kevin and Liu, Zicheng and Wang, Xinchao and Wang, Lijuan},
title = {MM-Vet: evaluating large multimodal models for integrated capabilities},
year = {2024},
publisher = {JMLR.org},
booktitle = {International Conference on Machine Learning},
articleno = {2381},
numpages = {25},
location = {Vienna, Austria},
}

@inproceedings{
  khan2024chitrarth,
  title={Chitrarth: Bridging Vision and Language for a Billion People},
  author={Shaharukh Khan and Ayush Tarun and Abhinav Ravi and Ali Faraz and Akshat Patidar and Praveen Kumar Pokala and Anagha Bhangare and Raja Kolla and Chandra Khatri and Shubham Agarwal},
  booktitle={NeurIPS Multimodal Algorithmic Reasoning},
  year={2024},
}

@InProceedings{Goyal_2017_CVPR,
  author    = {Goyal, Yash and Khot, Tejas and Summers-Stay, Douglas and Batra, Dhruv and Parikh, Devi},
  title     = {Making the V in VQA Matter: Elevating the Role of Image Understanding in Visual Question Answering},
  booktitle = {Proceedings of the IEEE Conference on Computer Vision and Pattern Recognition (CVPR)},
  year      = {2017},
  pages     = {6904--6913}
}

@misc{xai2024realworldqa,
  author = {xAI},
  title = {Grok-1.5 Vision Preview},
  howpublished = {\url{https://x.ai/blog/grok-1.5v}},
  month = {April},
  year = {2024}
}

@misc{hfrealworldqa,
  author = {xAI},
  title = {RealWorldQA Dataset},
  howpublished = {Hugging Face Dataset Repository},
  year = {2024},
  url = {https://huggingface.co/datasets/xai-org/RealworldQA}
}

@article{kj2025indicmmlu,
  title={IndicMMLU-Pro: Benchmarking Indic Large Language Models on Multi-Task Language Understanding},
  author={KJ, Sankalp and Kumar, Ashutosh and Balaji, Laxmaan and Kotecha, Nikunj and Jain, Vinija and Chadha, Aman and Bhaduri, Sreyoshi},
  journal={arXiv preprint arXiv:2501.15747},
  year={2025}
}

@article{wei2022chain,
  title={Chain-of-thought prompting elicits reasoning in large language models},
  author={Wei, Jason and Wang, Xuezhi and Schuurmans, Dale and Bosma, Maarten and Xia, Fei and Chi, Ed and Le, Quoc V and Zhou, Denny and others},
  journal={Advances in neural information processing systems},
  volume={35},
  pages={24824--24837},
  year={2022}
}

@article{cai2024temporalbench,
    title={TemporalBench: Towards Fine-grained Temporal Understanding for Multimodal Video Models},
    author={Cai, Mu and Tan, Reuben and Zhang, Jianrui and Zou, Bocheng and Zhang, Kai and Yao, Feng and Zhu, Fangrui and Gu, Jing and Zhong, Yiwu and Shang, Yuzhang and Dou, Yao and Park, Jaden and Gao, Jianfeng and Lee, Yong Jae and Yang, Jianwei},
    journal={arXiv preprint arXiv:2410.10818},
    year={2024}
  }

\clearpage

\appendix

\section{Appendix}

\label{sec:appendix}

\subsection{Qualitative Examples}
\begin{figure*}[ht] 
    \centering
    \includegraphics[width=1.05
    \textwidth]{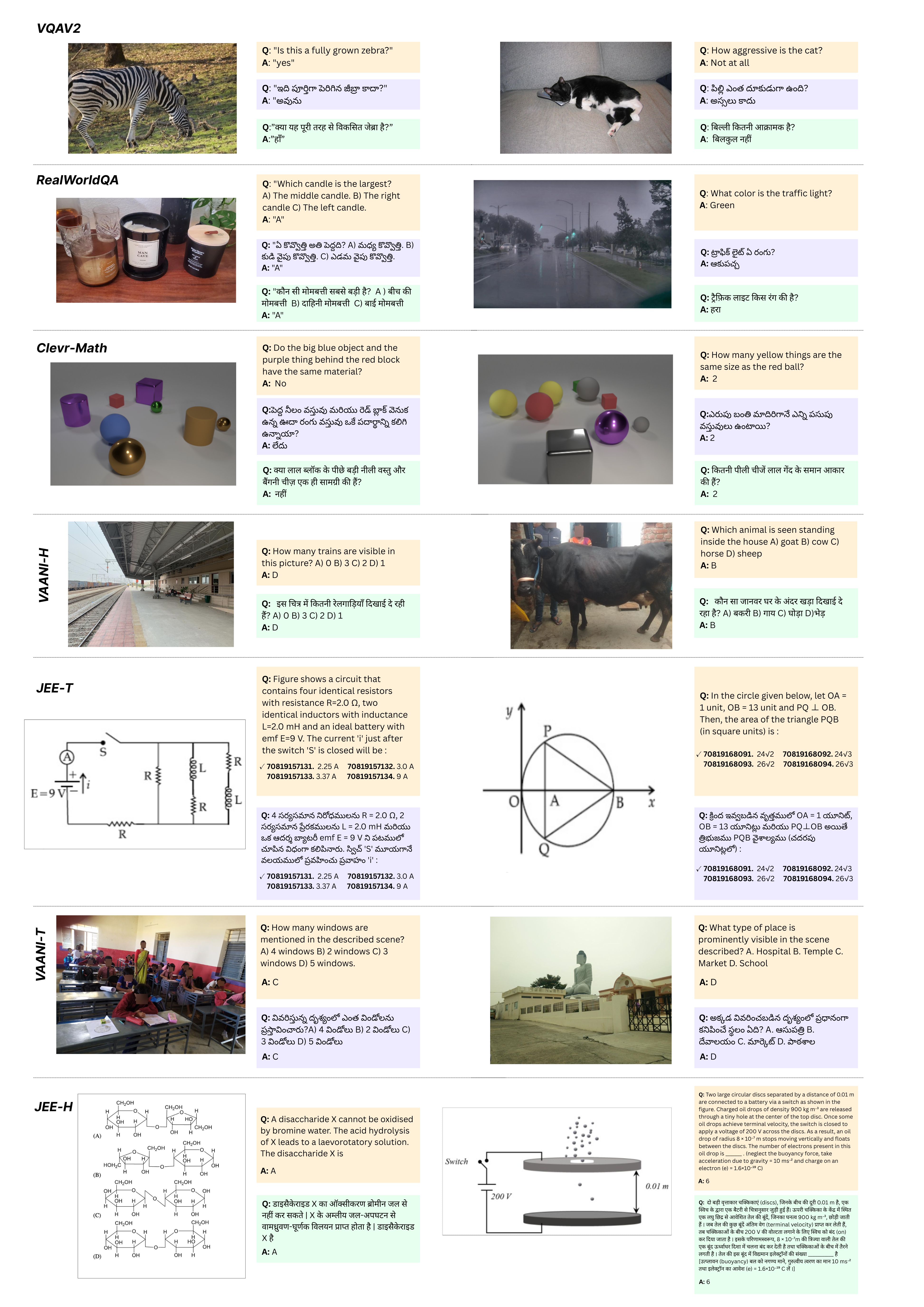} 
    \caption{Qualitative Examples from \textit{\textbf{VQAv2, RealWorldQA, CLEVR-Math, VAANI-H, JEE-T, VAANI-T, JEE-H}}}
    \label{fig:a1ds}
\end{figure*}

Refer to Fig. \ref{fig:a1ds}, with examples from each of the datasets, i.e \vqasource, RealWorldQA, \clevrmath, VAANI-H, JEE-T, VAANI-T, JEE-H

\subsection{Compute and Costs}
We primarily ran all open-source models on H100 and A100 GPUs, rented via the Akash Console Network\footnote{https://akash.network/}, incurring approximately 100 USD in compute costs. For proprietary models, we used APIs, leveraging Gemini’s free tier and spending around 60 USD on OpenAI’s APIs. The total expenditure amounts to approximately 160 USD.

\subsection{Prompts}

\label{app:prompts}
\subsection*{JEE-H COT Inference}
\begin{PromptBlock}
\textbf{eng_prompt} = """You are an expert at solving physics, chemistry and math questions that involve both text and images. Analyze the text and the associated image to answer the question above.
First, think step-by-step and provide your detailed reasoning. Explain how you interpret the text and the image, the principles or formulas you are using, and the intermediate calculations or logical steps you take to arrive at the solution.
After your reasoning, provide the final answer on a new line.
Your entire response, including the reasoning and the final answer, MUST end with the following line exactly:
\{ answer: \}"""
\end{PromptBlock}

\subsection*{JEE-H/T Standard Inference}
\begin{PromptBlock}
\textbf{eng_prompt} = "You are an expert at solving physics, chemistry and math questions that involve both text and images.
Analyze the text and the associated image to answer the question above. Provide only the final answer, without any explanations or intermediate steps. Your response MUST end with the following line: \{ answer: \}"
\end{PromptBlock}




\subsection*{VQA LLM Eval}
\begin{PromptBlock}
\textbf{System_Prompt} = "You are an expert judge evaluating semantic similarity between two answers.
Respond only with '1' for similar or '0' for not similar."

\textbf{User_Prompt} = "Are the following two answers semantically similar?

Answer 1: \{a\}
Answer 2: \{b\}

Respond with only '1' if they are similar in meaning, or '0' if they are not similar."
\end{PromptBlock}

\subsection*{CLEVR LLM Eval}
\begin{PromptBlock}
\textbf{System\_Prompt} = "You are an expert judge evaluating semantic similarity between two answers.
Respond only with '1' if they are semantically similar, or '0' if they are not."

\textbf{User\_Prompt} = "Are the following two \{lang\} answers semantically similar?

Answer 1: \{ans1\}  
Answer 2: \{ans2\}

Ignore minor differences in phrasing. Respond with '1' if they express the same meaning, or '0' otherwise."
\end{PromptBlock}

\subsection{Inference Strategies}
This appendix details the various inference techniques employed for evaluating the multimodal models, along with the specific prompt setups used for English and Hindi. Below are setups in English.



\subsection*{Caption-Only Baseline}
\label{app:socratic}
This technique involves a two-step process: \\
        \textbf{1) Dense Caption Generation:} The model first generates a detailed, factual description (dense caption) of the provided image. \\
        \textbf{2) Question Answering with Caption:} The model then answers the original question using \textit{only} the generated dense caption as context, without access to the original image.
    
Prompt for Step 1 (Dense Caption Generation):
\begin{PromptBlock} 
\textbf{eng_prompt} = Don’t forget these rules:
1. Be Direct and Concise: Provide straightforward descriptions without adding interpretative or speculative elements.
2. Use Segmented Details: Break down details about different elements of an image into distinct sentences, focusing on one aspect at a time.
3. Maintain a Descriptive Focus: Prioritize purely visible elements of the image, avoiding conclusions or inferences.
4. Follow a Logical Structure: Begin with the central figure or subject and expand outward, detailing its appearance before addressing the surrounding setting.
5. Avoid Juxtaposition: Do not use comparison or contrast language; keep the description purely factual.
6. Incorporate Specificity: Mention age, gender, race, and specific brands or notable features when present, and clearly identify the medium if it’s discernible.
When writing descriptions, prioritize clarity and direct observation over embellishment or interpretation. Write a detailed description of this image, do not forget about the texts on it if they exist. Also, do not forget to mention the type/style of the image. No bullet points. Start with the words, "This image displays:"
\end{PromptBlock}

Prompt for Step 2 (Question Answering with Caption):
        \begin{PromptBlock}
        Context: \{generated_caption\}
        Question: \{question_text\}
        Answer: 
        \end{PromptBlock}

\subsection*{Chain of Thought (CoT) Prompting}
\label{app:cot}
This technique guides the model to generate a step-by-step thought process (rationale) before arriving at the final answer. This is intended to improve reasoning capabilities for complex questions.

The image is provided along with language-specific system and user prompts.
    
            \begin{PromptBlock}
\textbf{English System Prompt:} 
    When provided with an image and a question, generate a rationale first and then derive an answer.
    Your rationale should include detailed visual elements in order to derive the answer.
            \end{PromptBlock}
            \begin{PromptBlock}
\textbf{English User Prompt:}  
    Answer the question with following instruction: 
    1. Generate a rationale first and then derive an answer. 
    2. For your final answer, provide a Correct Option Number Only. 
    Question: 
    \{question\} 
    \# Output Format \# 
    <rationale> 
    \#\#\# Answer: <your answer>
        \end{PromptBlock}

\subsection*{Standard (Direct Inference)}
\label{app:direct_inference}
The model is provided with both the image and the question and is expected to generate a direct answer without explicit intermediate reasoning steps requested in the prompt. \\
The image and the question are provided to the model. The prompt typically instructs the model to answer directly, often in a specific format (e.g., option number for multiple-choice questions).

Example input (for a multiple-choice question):
\begin{PromptBlock}
    
\textbf{Image:} [Image Data]
Question: \{question_text_with_options\}
Instruction: Provide the Correct Option Number Only.
Answer:
\end{PromptBlock}
        (Note: This often shares the final answer formatting instruction with CoT, but without the explicit rationale generation step.)

\subsection*{JEE-H Scoring Rules}
\label{app:JEE-H score}
\begin{itemize}
    \item For single correct Multi-Choice Questions (MCQs) and integer-type questions, a binary score was assigned: 1 if the model answered with the correct option/integer and 0 otherwise.
    \item For multi-correct MCQs, a score of 1 is given when the generated response consists of all and only the correct options. If any of the wrong option is chosen, the response is given a score of 0. If the response contains no incorrect option and some of the correct options, a score of 0.25 is given for each of the correct options.
    \item For numeric-type questions, answers in the range of ±0.01 with the gold answer are given a score of 1, and 0 otherwise.
    \item In Chain-of-Thought (CoT) inferencing, if a Hindi response included reasoning in English, 0 was given irrespective of the answer.
    \item In standard zero-shot inference, where the model was explicitly instructed to return only a direct answer, responses containing any explanation or reasoning were also scored 0, irrespective of the answer.
\end{itemize}

\subsection*{Impact of LLM Judge on Evaluation Scores}
\label{app:inc_hybrid_eval}
As detailed in Section \ref{section: eval_metrics}, our hybrid evaluation strategy for CLEVR-Math and VQA incorporates an LLM judge to re-evaluate answers initially marked incorrect by the exact match protocol. This approach revealed a notable trend: the score increase attributed to the LLM judge was considerably more for the responses in Hindi and Telugu as compared to English.
For instance, when evaluating the VQA dataset, the average scores for English responses increased by $\approx 4\%$ after LLM judging. In contrast, Hindi scores on the same dataset increased by $\approx 12\%$ after the initial exact match scores. This disparity suggests that the models demonstrate a greater proficiency in adhering to strict answer formatting conventions in English, while responses in Hindi and Telugu, though often semantically correct, are more frequently penalized by the exact match criteria due to variations in phrasing or formatting. The LLM judge effectively mitigates this by recognizing a broader range of correct expressions.

\subsection{Licenses}
The datasets used in our benchmark are released under the following licenses:
\begin{itemize}
    \item VQAv2, VAANI and ClevrMath are released under Creative Commons Attribution 4.0 International (CC BY 4.0) license.
    \item RealWorldQA is released under Creative Commons Attribution No Derivatives 4.0.
    \item JEE-Vision is built from publicly available JEE questions. We will be releasing links under CC BY-NC 4.0 license along with the QA pair. The exam papers are distributed for the general public (https://www.jeeadv.ac.in/archive.html) used by various publishers and institutes for commercial purposes already. When collecting this data, we followed the steps by previous works on JEE Bench~\cite{arora-etal-2023-llms-jeebench}.
\end{itemize}
    
\end{document}